\PassOptionsToPackage{dvipsnames}{xcolor}
\PassOptionsToPackage{breaklinks}{hyperref}

\documentclass{arxiv}

\usepackage{booktabs}
\usepackage{todonotes}
\usepackage{subfig}
\usepackage{amsfonts}
\usepackage{bbm}
\usepackage{subcaption}
\usepackage{multirow}
\usepackage{threeparttable}
\newcommand{\gain}[1]{\textcolor{ForestGreen}{\textbf{+#1}}}
\newcommand{\tcite}[1]{{\hypersetup{citecolor=black}\cite{#1}}}
\newcommand{\relgain}[1]{\scriptsize\textcolor{gray}{(#1\%)}}
\usepackage{xspace}
\newcommand{\eg}{e.g.\@\xspace}

\newcommand{\vs}{vs.\@\xspace}

\usepackage{color}
\usepackage{amsmath,amssymb}

\begin{document}

\title{EventCLR: Contrastive Self-Supervised Learning for Few-Shot Event-Based Vision using Spiking Neural Networks}

\author{Maxime Vaillant$^{1,2,4,*}$\orcid{0009-0006-5948-948X}, Axel Carlier$^{1,3}$\orcid{0000-0002-6838-3445}, Lai Xing Ng$^{1,4}$\orcid{0000-0002-5457-6289}, Christophe Hurter$^{1,3}$\orcid{0000-0003-4318-6717} and Benoit R. Cottereau$^{1,5}$\orcid{0000-0002-2624-7680}}

\affil{$^1$CNRS, IPAL IRL 2955, Singapore}
\affil{$^2$Université de Toulouse, IRIT, France}
\affil{$^3$Fédération ENAC ISAE-SUPAERO ONERA, Université de Toulouse, France}
\affil{$^4$Institute for Infocomm Research, A*STAR, Singapore}
\affil{$^5$CerCo, CNRS UMR 5549, Université de Toulouse, France}
\affil{\par\vspace{2pt}$^*$Author to whom any correspondence should be addressed.}

\email{maxime.vaillant@utoulouse.fr}


\begin{abstract}Event-based vision sensors provide significant advantages for high-speed perception, including microsecond temporal resolution, high dynamic range, and low power consumption. When combined with Spiking Neural Networks (SNNs), they are well suited to neuromorphic hardware and energy-efficient applications on embedded systems. However, this potential is severely limited by the scarcity of large-scale labeled datasets required to effectively train such models. In this work, we present \textbf{EventCLR}, a systematic approach to contrastive self-supervised learning for SNNs under label scarcity, a setting that closely matches the practical constraints of most real-world neuromorphic sensing applications. We instantiate this approach with a standard SimCLR-style pipeline adapted to the spiking domain via surrogate gradient training, deliberately keeping the architecture simple, and pair it with a systematic ablation of event-specific augmentation families (spatial, temporal, and polarity transformations). Through extensive experiments on different event datasets (CIFAR10-DVS, N-Caltech101, N-MNIST, DVS-Gesture, and N-Cars), we demonstrate that self-supervised pretraining with subsequent fine-tuning outperforms supervised learning in low-data regimes, achieving consistent gains in few-shot and semi-supervised settings. Our ablation studies show that the combination of the three families works best, and that spatial and temporal augmentations are the most complementary pair. Unlike concurrent work that focuses on large-scale spiking self-supervised pretraining through architectural innovation, EventCLR directly addresses the problem of label efficiency by providing a controlled benchmark for the low-label regime, which remains largely unexplored by existing large-scale approaches. We further show that learned representations transfer across datasets, contributing to efforts for powerful event-based models in label-scarce settings.
\end{abstract}

\section{Introduction}

Event-based vision sensors, or Dynamic Vision Sensors (DVS), depart from frame-based imaging by measuring per-pixel changes in log-intensity asynchronously, producing sparse streams of events rather than full frames at fixed intervals. This yields microsecond-level temporal resolution, high dynamic range ($>$120~dB), and low power consumption, making event cameras well-suited for fast motion, challenging lighting, and resource-constrained platforms.

\begin{figure*}[t]
    \centering
    \includegraphics[width=\linewidth]{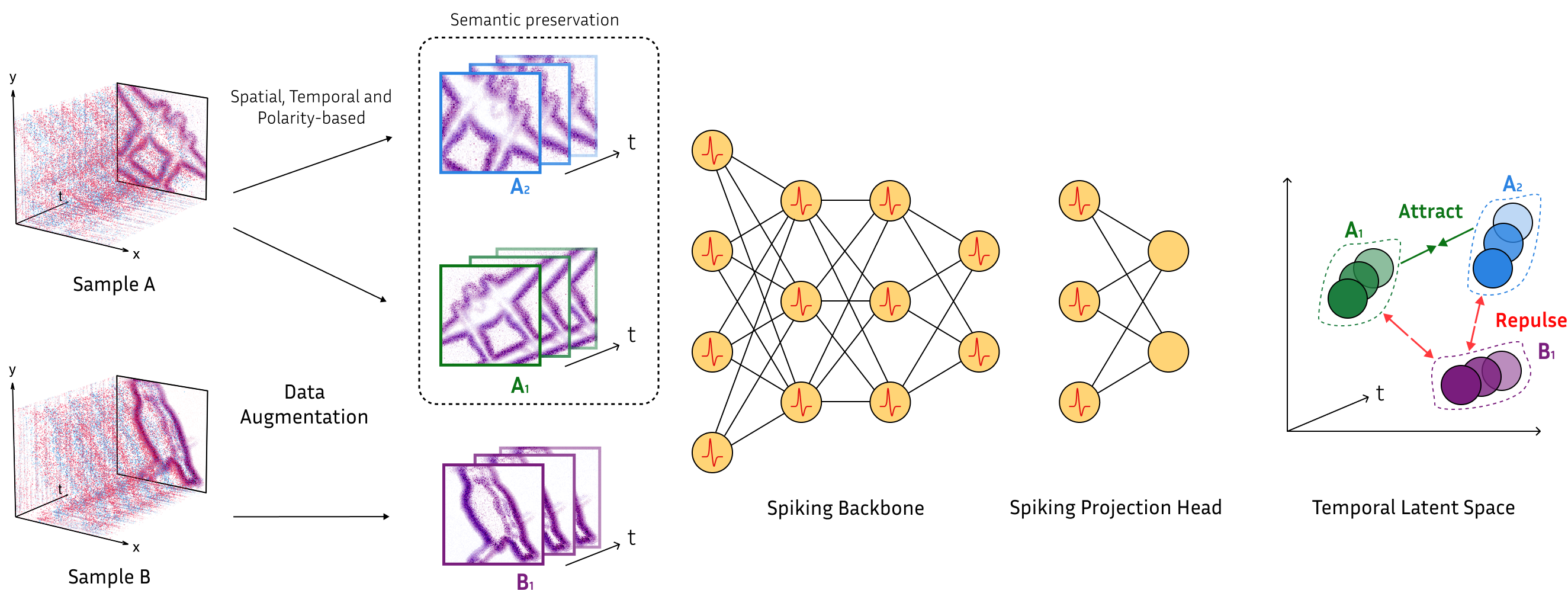}
    \caption{Overview of the EventCLR pipeline. Two augmented views ($A_1$, $A_2$) are generated from the same raw event stream and form a positive pair, while a third view ($B_1$), generated from a different sample, acts as a negative. Each view is processed independently by the shared spiking backbone $f(\cdot)$ over $T$ discrete timesteps, and the spiking projection head $g(\cdot)$ maps the resulting spike trains to a non-spiking, real-valued embedding at every timestep, producing a temporal latent space $\{z_v[t]\}_{t=1}^{T}$ per view (time-averaged before the NT-Xent contrastive loss, Eq.~\ref{eq:ntxent}). In the resulting embedding space (right), the NT-Xent objective pulls the positive pair $A_1, A_2$ together (\emph{Attract}) while pushing both away from the negative $B_1$ (\emph{Repulse}).}
    \label{fig:pipeline_overview}
\end{figure*}

The asynchronous, spike-based nature of event data aligns with Spiking Neural Networks (SNNs)~\cite{snn}, which process information as discrete spike sequences. Because SNNs explicitly model time through spiking dynamics, they are a natural fit for event streams: each neuron integrates incoming currents into its membrane potential and emits a spike when a threshold is exceeded, enabling sparse and energy-efficient processing.

Despite these advantages, deploying SNNs for event-based vision faces a critical bottleneck: the scarcity of large-scale labeled datasets. Event streams are not human-readable, so annotation requires specialist tooling and domain expertise; current benchmarks remain orders of magnitude smaller than conventional image datasets (e.g., N-MNIST: 60k, CIFAR10-DVS: 10k vs ImageNet: 1.2M), and most are constructed by re-recording static images with moving cameras rather than by capturing naturally dynamic scenes. Models therefore often learn from synthetic motion patterns that only partially reflect deployment conditions, making label-efficient training especially important.

Event cameras are often considered together with neuromorphic hardware (e.g., Intel Loihi, SpiNNaker) for applications such as autonomous inspection, driving, or embedded robotics, and in these applications labeled data are hard to obtain. For example, a drone inspecting a hard-to-access site for anomalies will only see a few anomalies, and annotating them requires domain expertise. In such cases, pretraining on \emph{unlabeled} event streams and then adapting the model with a few labeled examples is the realistic setting. This deployment context motivates our study, while energy consumption and execution on neuromorphic hardware are beyond the scope of this work.

Self-supervised learning (SSL) has emerged as a powerful paradigm to learn visual representations without labels by exploiting the structure inherent in unlabeled data. Contrastive methods such as SimCLR~\cite{simclr} have achieved strong results in conventional vision by learning invariances through data augmentation. However, their application to event-based data and SNNs remains largely unexplored. Key challenges include: (i) designing meaningful augmentations for sparse, polarity-based event representations, and (ii) training SNNs with contrastive objectives in the presence of discrete, non-differentiable spike operations.

Beyond standard linear evaluation, label scarcity is also evident in few-shot scenarios: for many event datasets, only a handful of labeled recordings per class are available when adapting to a new sensor, domain, or acquisition protocol. Few-shot evaluation therefore provides a practical stress test for representation quality: a good encoder should separate classes with a small classifier trained on $k\in\{1, 10, 20, 50\}$ examples per class, rather than relying on large supervised fine-tuning sets. To study this label scarcity, we present \textbf{EventCLR}, a contrastive pretraining approach for SNNs on event data, which adapts SimCLR to the spiking domain with surrogate-gradient learning (Figure~\ref{fig:pipeline_overview}), and we evaluate how pretrained features become informative in few-shot and semi-supervised settings. EventCLR takes event data as input, but events are binned into $T = 4$ histograms before the network and the loss is computed on time-averaged embeddings, so individual events play little role. The pipeline is closer to a stateless method than to a stateful one (Section~\ref{sec:contrastive}). Its main contribution is the adaptation of SimCLR to event-based vision datasets, with a recipe to augment the input data, demonstrated on a time-stepped SNN. Our main contributions are:
\begin{itemize}
    \item A controlled low-label evaluation of contrastive SSL for SNNs: few-shot ($k\in\{1, 10, 20, 50\}$) and label-fraction evaluation on five event datasets, covering static-image-derived and dynamic recordings, with in-domain and cross-dataset pretraining.
    \item A pre-/post-representation augmentation pipeline with three families of event augmentations (spatial, temporal, and polarity). The pipeline separates transformations applied to the raw event stream from those applied to the grid representation.
    \item Evidence on CIFAR10-DVS that, for contrastive pretraining of SNNs, combining spatial, temporal, and polarity augmentations gives the best few-shot accuracy, with spatial and temporal as the most complementary pair, and that augmentation sets designed for supervised training (EventDrop-style, NDA-style) do not reach this level.
\end{itemize}

Overall, in the 1-shot regime, EventCLR improves fine-tuning accuracy over supervised training from scratch by +7.7 points on CIFAR10-DVS, +13.4 on N-Caltech101, +15.1 on DVS-Gesture, +6.1 on N-Cars and +39.3 on N-MNIST (Table~\ref{tab:main_results}), and the gains remain as more labeled data becomes available. Code will be available upon publication.

\section{Related Work}

\subsection{Contrastive Learning for Frame-Based Vision}

SSL has become a dominant paradigm for learning transferable visual features without labels. Contrastive approaches \cite{contrastive1, contrastive2, swav, contrastive3} such as SimCLR \cite{simclr} or MoCo \cite{moco} learn representations by maximizing agreement between different augmented views of the same sample, while simultaneously contrasting them against representations of other samples in the batch, which act as negative examples to prevent representational collapse. Non-contrastive methods such as BYOL \cite{byol} or SimSiam \cite{simsiam} remove explicit negatives and prevent collapse through architectural asymmetry and optimization constraints. These methods have demonstrated impressive results on ImageNet and other large-scale RGB datasets, achieving performance competitive with supervised pretraining when sufficient unlabeled data is available. However, their direct application to event-based data and spiking architectures remains largely unexplored.

\subsection{Augmentation Strategies for Event Data}

\begin{figure*}[t]
    \centering
    \subfloat[Original]{
        \includegraphics[width=0.20\linewidth]{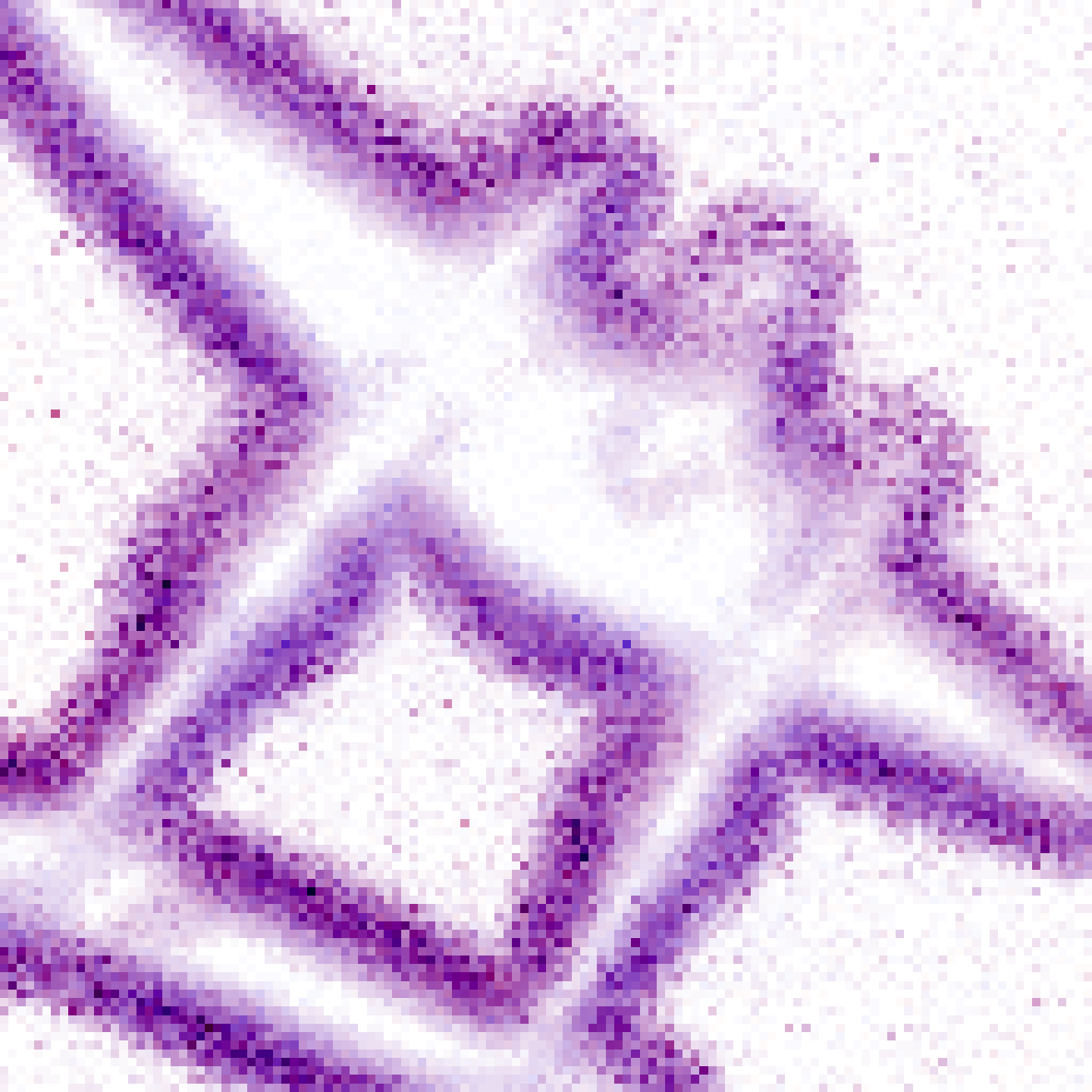}
        \label{fig:aug_original}
    }
    \subfloat[Spatial]{
        \includegraphics[width=0.20\linewidth]{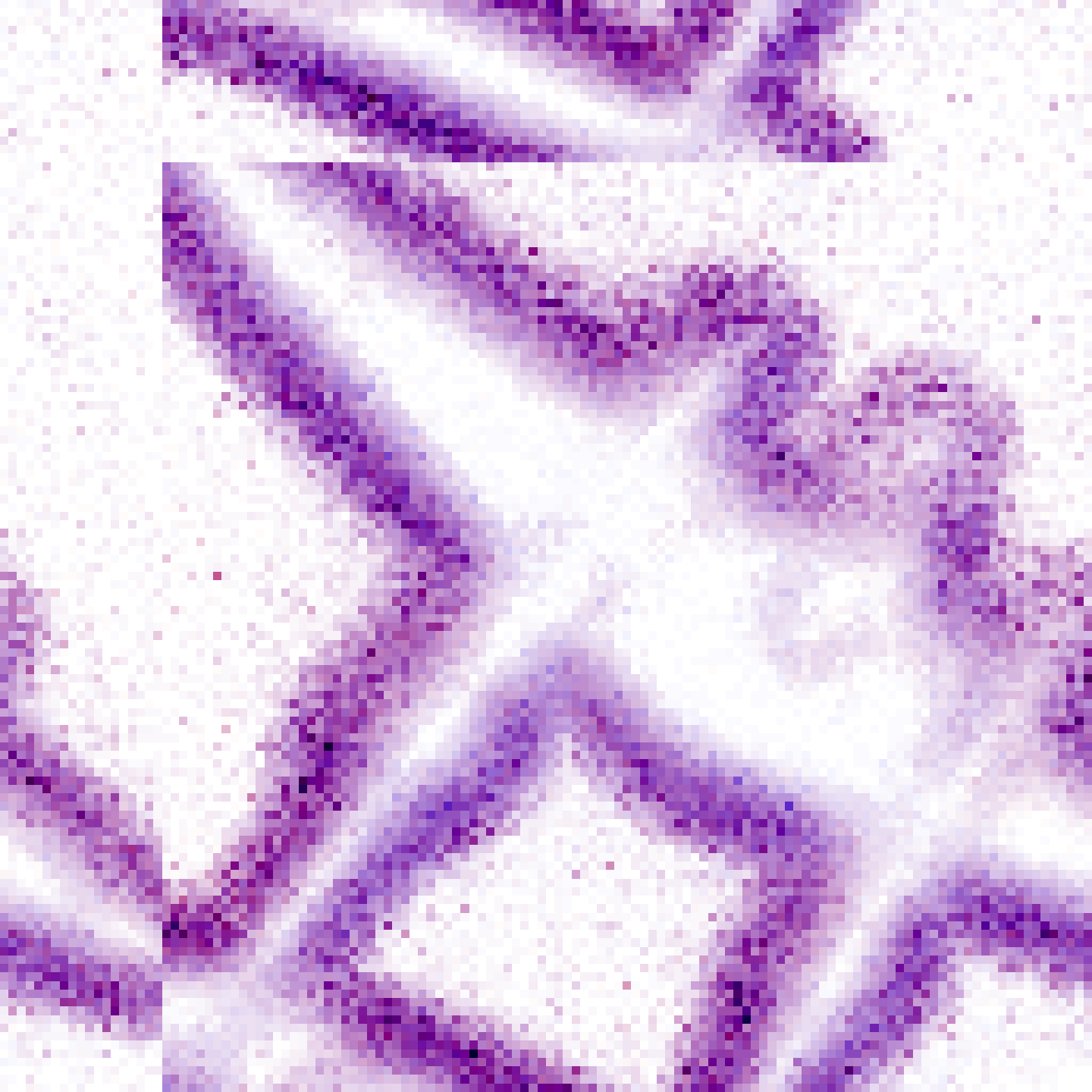}
        \label{fig:aug_spatial}
    }
    \subfloat[Polarity]{
        \includegraphics[width=0.20\linewidth]{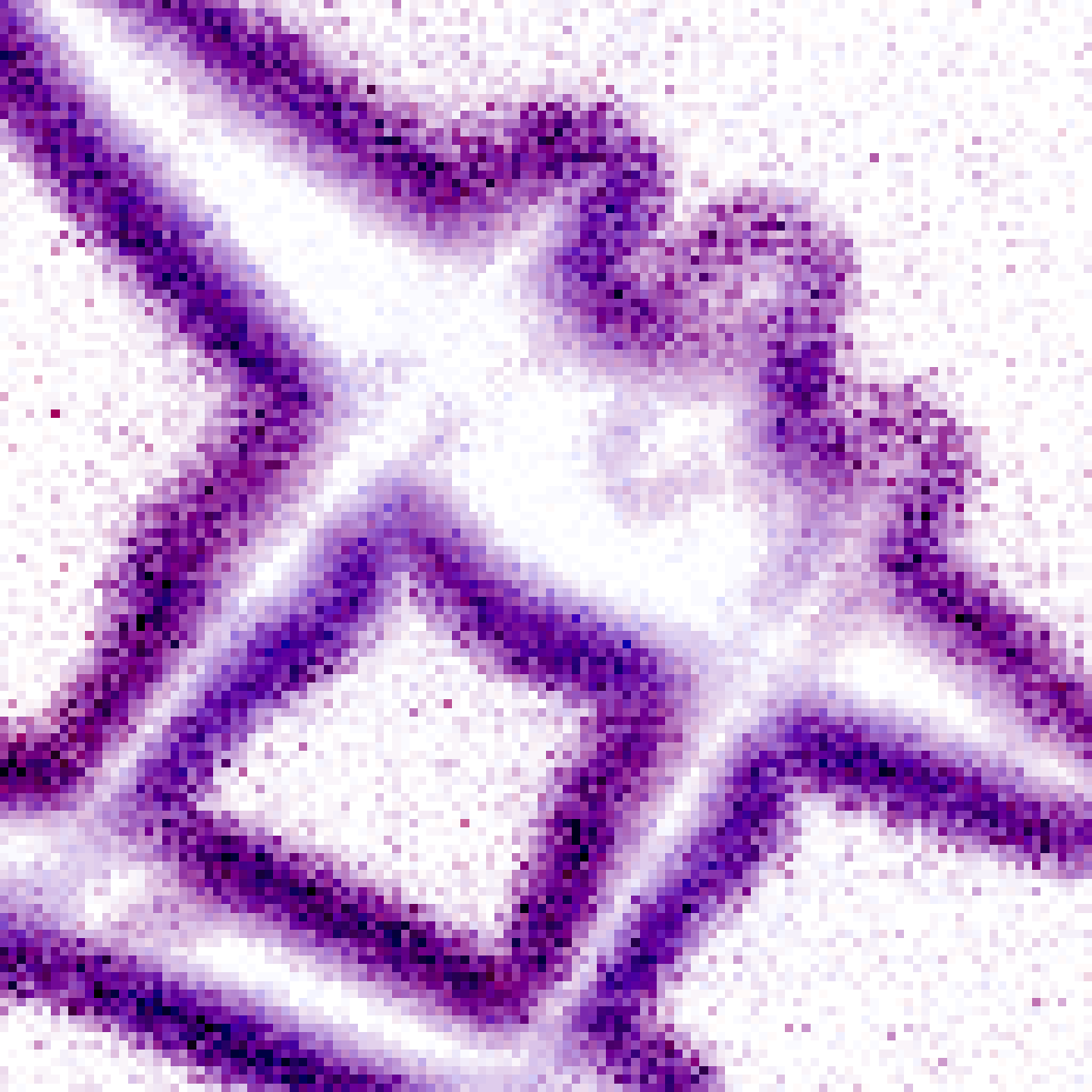}
        \label{fig:aug_pol_bright}
        \label{fig:aug_polarity}
    }
    \subfloat[Temporal]{
        \includegraphics[width=0.20\linewidth]{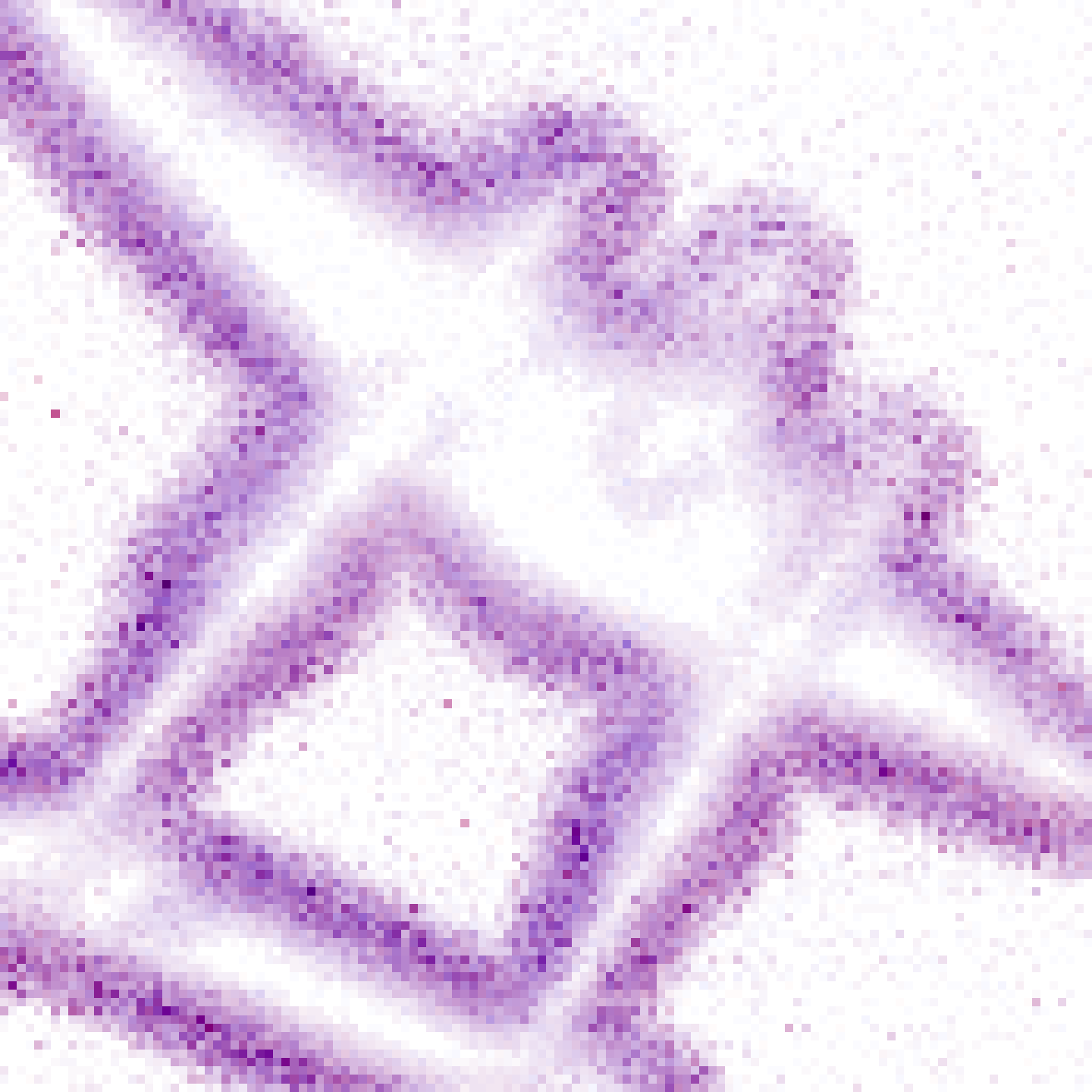}
        \label{fig:aug_temporal}
    }
    \caption{Effect of each augmentation family on an event histogram from CIFAR10-DVS. (a)~Original sample. (b)~Spatial: rolling shifts the content spatially. (c)~Polarity: brightness scaling modifies event intensities. (d)~Temporal: cropping a different time window changes the motion phase. More examples are provided in the Supplementary Material.}
    \label{fig:augmentations}
\end{figure*}

Data augmentation is a critical component of contrastive learning~\cite{simclr, aug1}, as it determines the invariance encoded in the learned representations. While augmentation strategies for frame-based vision are well established, their direct application to sparse, asynchronous event streams is challenging. Early work~\cite{aug2} primarily focused on supervised learning, with EventDrop~\cite{eventdrop} introducing temporal augmentations such as random event dropping or spatio-temporal removal. Other methods like EventMix~\cite{eventmix} and EventAugment~\cite{eventaugment} explored mixing strategies and learned policies. However, augmentations designed for supervised robustness are not necessarily optimal for contrastive learning~\cite{simclr}, which aims for broader invariance rather than specific robustness. In this work, we systematically evaluate event-specific spatial, temporal, and polarity augmentations to identify the most effective combination for self-supervised learning, demonstrating that a carefully designed strategy is key to learning good representations.

\subsection{Self-Supervised Learning for SNNs and Event Data}
\label{sec:related_work_ssl_snn}

Self-supervised learning for spiking neural networks and event data is still in its early stages. A first line of work tackles the problem through multi-modal fusion, aligning event streams with conventional RGB frames. For instance, \cite{sslsnn} trains an SNN encoder that jointly embeds paired frame-event inputs, while \cite{eventpretraining} follows a similar cross-modal alignment strategy using a standard ANN backbone. Although both demonstrate improved representations, they fundamentally depend on the availability of synchronized frame-event datasets, which limits their applicability when only event data is available.

Closer to our setting, several recent methods investigate SSL using event data alone. Neuromorphic Data Augmentation (NDA)~\cite{nda} and NeuroMOCO~\cite{neuromoco} adapt established SSL frameworks to SNNs (SimSiam and MoCo, respectively) and report promising results in high-data regimes. However, the available code and implementation details are insufficient to fully reproduce the reported results. Similarly, \cite{jepasnn} explores a non-contrastive objective based on VICReg~\cite{vicreg}, eliminating the need for negative pairs via variance, invariance, and covariance regularization. With a non-spiking backbone, Masked Event Modeling (MEM)~\cite{mem} adapts BEiT-style masked token prediction to event histograms and pretrains a ViT without labels or RGB images. It improves classification on N-ImageNet, N-Caltech101, and N-Cars, and also shows that a larger pretraining set does not always help: pretraining on N-ImageNet improves N-Caltech101, but on N-Cars in-domain pretraining stays better (98.6\% \vs 93.3\%). MEM evaluates label efficiency with fractions of 10\% or more of the training set, whereas we study spiking backbones with 1 to 50 labels per class. Despite differences in objective functions, these works share a common limitation: they do not investigate the few-shot (low-label) regime. More recently, SpikeCLR~\cite{spikeclr} tackles a complementary challenge, scaling fully self-supervised SNN training from small neuromorphic benchmarks to ImageNet-1K by introducing a dual-path MixedLIF neuron together with temporal alignment losses that stabilize surrogate-gradient training across augmented views. It reports strong results, matching or outperforming non-spiking SSL baselines on large-scale pretraining and downstream COCO detection and segmentation tasks. However, at the time of writing, the publicly available resources for SpikeCLR, as for NDA and NeuroMOCO, were not sufficient to support a faithful reproduction of the method. Consequently, we were unable to include it as a reproducible baseline in this study. In contrast, EventCLR provides a systematic study of SSL pretraining for SNNs under label scarcity, explicitly quantifying improvements in data efficiency and generalization.

\section{Methods}
\label{sec:methods}

\subsection{Event-Based Vision and Representations}
Event-based cameras \cite{eventsurvey} output a stream of events of the form $e=\langle t, x, y, p \rangle$, where $t$ denotes the timestamp, $(x,y)$ the pixel location, and $p \in \{-1,+1\}$ the polarity of the brightness change \cite{eventcamera}. Their asynchronous nature enables microsecond-level temporal resolution and high dynamic range, which has motivated applications in recognition, tracking, and high-speed perception. For learning-based pipelines, event streams are often converted into grid-based representations (e.g., accumulated event frames \cite{eventhistogram}, voxel grids \cite{voxelgrid, voxelgrid2} or time surfaces\cite{timesurface, timesurface2, timesurface3}) to leverage convolutional architectures while retaining key spatio-temporal information. In this work, we use a sequence of normalized event histograms as input. Specifically, we partition the time span $[t_0, t_{\text{end}}]$ of the event stream into $T$ consecutive bins of equal duration, and count the events of each polarity at each pixel within each bin:
\begin{equation}
\label{eq:histogram}
    \mathcal{H}[n, c, y, x] = \sum_{k} \mathbbm{1}\left[n_k = n,\ c_k = c,\ (y_k, x_k) = (y, x)\right], \qquad n_k = \left\lfloor T\,\frac{t_k - t_0}{t_{\text{end}} - t_0} \right\rfloor,
\end{equation}
where $c_k \in \{0, 1\}$ indexes the polarity of event $e_k$. The counts are then min--max normalized over the whole tensor, $\bar{\mathcal{H}} = (\mathcal{H} - \min \mathcal{H})/(\max \mathcal{H} - \min \mathcal{H})$, and each map is resized to a common resolution with adaptive max pooling. We use $T = 4$ and $H \times W = 48 \times 48$ for all datasets. This yields a spatio-temporal tensor $\mathbf{x} \in \mathbb{R}^{T \times 2 \times H \times W}$, which keeps a coarse temporal structure ($T = 4$ bins) and lets the SNN integrate information over these time steps. We use event histograms because they are a simple and widely used representation. Supplementary Section~S3.5 compares them with non-normalized histograms and voxel grids, with similar results.

\subsection{Spiking Neural Networks}

Spiking Neural Networks (SNNs)~\cite{snn2, snn3, snn4, snn1} process information using discrete events (spikes) rather than continuous values, closely mimicking biological neural processing. This event-driven nature makes them naturally suited for processing asynchronous data streams from event cameras. In this work, we employ the Leaky Integrate-and-Fire (LIF)~\cite{lif1, lif2} neuron model. The dynamics of a LIF neuron $i$ at layer $l$ are governed by its membrane potential $u_i^{(l)}(t)$. In the discrete time domain, the update rule is given by:
\begin{equation}
    \label{eq:lif}
    u_i^{(l)}[t] = \beta\, u_i^{(l)}[t-1] \big(1 - s_i^{(l)}[t-1]\big) + \sum_j w_{ij} s_j^{(l-1)}[t],
\end{equation}
where $\beta \in (0,1)$ is the decay factor representing the membrane leak, $w_{ij}$ are the synaptic weights connecting neuron $j$ in the previous layer to neuron $i$, and $s_j^{(l-1)}[t] \in \{0, 1\}$ represents the input spike. A neuron emits a spike $s_i^{(l)}[t]$ when its membrane potential exceeds a firing threshold $V_{th}$:
\begin{equation}
    s_i^{(l)}[t] = \Theta(u_i^{(l)}[t] - V_{th}),
\end{equation}
where $\Theta(\cdot)$ is the Heaviside step function. After firing, the membrane potential is reset to zero (hard reset): the factor $(1 - s_i^{(l)}[t-1])$ removes the potential accumulated before the spike, so the neuron restarts from its input alone. We use the same parameters for all layers: $\beta = 0.5$ (membrane time constant of 2 timesteps), fixed and not learned, and $V_{th} = 1$. The input current is not scaled by the leak, as in Eq.~(\ref{eq:lif}). Membrane potentials are set to zero at the start of each sample.

\paragraph{Training.}
The non-differentiable nature of the spike generation function prevents standard backpropagation. To address this issue, we use the surrogate gradient method \cite{surrogate} during the backward pass. We approximate the gradient of the Heaviside function with the derivative of an arctan function (ATan surrogate of SpikingJelly~\cite{spikingjelly}), allowing gradients to flow through the spiking layers to update synaptic weights. In practice, we unroll the SNN dynamics over time and optimize the resulting computation graph with Backpropagation Through Time (BPTT).

\paragraph{Backbone.} We adopt SEW-ResNet-18~\cite{sewresnet} as our main SNN backbone. This model adapts ResNet-18 to the spiking domain via Spiking Element-Wise (SEW) operations and residual blocks built from spiking neurons. ResNet-18 is a widely used architecture in conventional (frame-based) vision, making it a natural reference point for our spiking encoder. We use the ADD connection function of SEW. Convolution weights use Kaiming normal initialization, BatchNorm weights and biases are initialized to 1 and 0, and the last BatchNorm of each residual block is initialized to zero. In addition, we evaluate a separable-convolution \cite{convsep} lightweight variant and a Spiking VGG9 backbone to check that our conclusions are not architecture-specific.

\subsection{Datasets}
\label{sec:datasets}

We consider two kinds of neuromorphic datasets: (i) \emph{static-image-derived} datasets obtained by re-recording conventional image datasets with a moving event camera (re-display setup), and (ii) \emph{dynamic event-stream} datasets that capture naturally time-varying motions and actions. Our main experiments are conducted on CIFAR10-DVS and also on DVS-Gesture, covering both regimes. We additionally report results on N-Caltech101, N-MNIST, and N-Cars for broader comparison.

\paragraph{Static-Image-Derived Datasets.}
CIFAR10-DVS~\cite{cifar10dvs}, N-Caltech101~\cite{nmnist}, and N-MNIST~\cite{nmnist} follow a re-display protocol, where events are induced by controlled camera motion over static images. For CIFAR10-DVS in particular, samples are commonly recorded with a DVS128/DAVIS sensor at \(128\times128\) resolution while the camera performs predefined motion patterns in front of a monitor, yielding short event sequences per image and class. It contains approximately 10,000 event sequences across 10 object classes. As no official split is provided, we apply a 90/10 train/validation split for our experiments. N-Caltech101 and N-MNIST are both obtained from their respective frame-based counterparts (Caltech101 and MNIST) using the same acquisition protocol: an ATIS sensor executes three controlled microsaccades in front of a monitor displaying each static image, converting it into a short event stream. N-Caltech101 defines a 101-way object classification task over 8,709 samples inherited from Caltech101's original, variable-resolution images and class distribution; the 101 classes are the 100 object categories and the \texttt{BACKGROUND\_Google} class. As with CIFAR10-DVS, no official split is provided, so we again apply a stratified 90/10 train/validation split. Classes are unbalanced, and after the split (7,795 training and 914 test samples) the smallest class has 27 training samples. Labeled subsets are sampled without replacement and no class is excluded: when a class has fewer than $k$ training samples, we use all of them, which at $k = 50$ concerns 45 of the 101 classes. N-MNIST defines a 10-way digit classification task at a fixed $34\times34$ resolution, following the official MNIST split of 60,000 training and 10,000 test samples. Compared to CIFAR10-DVS, N-Caltech101 is substantially more challenging because of its much larger number of classes, whereas N-MNIST is considerably easier due to its simple, low-variability digit shapes. Together, these datasets span a broad range of task difficulty within the static-image-derived event benchmark regime. For pretraining only, we also use N-ImageNet~\cite{nimagenet}, recorded with a $640\times480$ event camera moving in front of a monitor displaying ImageNet-1K images. We use its training split without labels, with the same $48\times48$ representation as the other datasets.

\paragraph{Dynamic Event-Stream Datasets.}
DVS-Gesture~\cite{dvsgesture} is recorded with a DVS sensor (\(128\times128\) pixels) while subjects perform hand/arm gestures in front of the camera. Compared to re-display datasets, the events here are generated by real motion and temporal structure (gesture execution), which makes it a good testbed for spatio-temporal representations.
It contains gestures performed by 29 subjects. We use the standard subject-disjoint split: subjects 1--23 for training and 24--29 for testing. We load it through Tonic~\cite{tonic}, which gives 1,077 training and 264 test samples over 11 classes, including the ``other'' class. We additionally consider N-Cars~\cite{timesurface2}, recorded with an ATIS sensor mounted behind the windshield of a car during real urban driving, and defining a binary classification task (car \vs background) over 24,029 samples (15,422 train / 8,607 test). Unlike DVS-Gesture, where the event dynamics are driven primarily by the motion of the subject, the event dynamics in N-Cars are dominated by the ego-motion of the recording vehicle. This provides a complementary test of generalization to naturally dynamic, real-world event streams acquired with a different sensor. Detailed few-shot results on N-Cars are provided in the Supplementary Material.

\subsection{Contrastive Learning}
\label{sec:contrastive}

Our approach builds on SimCLR~\cite{simclr}, a contrastive self-supervised learning framework that learns visual representations by bringing together different augmented views of the same sample while pushing apart views coming from different samples. In our setting, each event recording is first converted into a grid-based representation (event histogram), on which we apply stochastic augmentations to generate two correlated views. Figure~\ref{fig:pipeline_overview} summarizes the resulting EventCLR pipeline, from raw event stream to the temporal latent space used for the contrastive loss.

\paragraph{Architecture.}
Given an input representation $x$, we sample two augmentations to obtain $x_1$ and $x_2$. Both views are processed by an encoder $f(\cdot)$ (SEW-ResNet-18 in our experiments), producing features $h_1=f(x_1)$ and $h_2=f(x_2)$. We append a small spiking projection head $g(\cdot)$ to map features to a contrastive space: $z_1=g(h_1)$ and $z_2=g(h_2)$. Similarity is computed in this space using cosine similarity after $\ell_2$ normalization.

\paragraph{Contrastive Objective (NT-Xent).}
Training is performed on mini-batches of size $N$. Each sample yields two views, resulting in $2N$ embeddings. For an anchor embedding $z_i$, its positive $z_j$ corresponds to the other view of the same original sample, while the remaining $2N-2$ embeddings act as negatives. The NT-Xent loss is defined as:
\begin{equation}
\label{eq:ntxent}
\ell_{i,j}=- \log\frac{\exp\left(\mathrm{sim}(\mathbf{z}_i, \mathbf{z}_j) / \tau \right)}
{\sum_{k=1}^{2N}\mathbbm{1}_{[k \neq i]}\exp\left(\mathrm{sim}(\mathbf{z}_i, \mathbf{z}_k) / \tau \right)} ,
\end{equation}
where $\mathrm{sim}(\cdot,\cdot)$ denotes cosine similarity and $\tau$ is a temperature hyperparameter. The final objective averages this loss over all positive pairs in the batch.

\paragraph{Temporal Aspect.}
Unlike standard SimCLR with ANN backbones, our spiking encoder is run for $T$ discrete timesteps and produces a sequence of representations. Concretely, for each view $x_v$ ($v\in\{1,2\}$) we obtain time-indexed embeddings $\{z_v[t]\}_{t=1}^{T}$ (a \emph{temporal batch} of size $T$). To use the usual batch-wise NT-Xent objective, we collapse this temporal dimension into a single vector per view by time-averaging: $\bar{z}_v = \frac{1}{T}\sum_{t=1}^{T} z_v[t]$. The contrastive loss in Eq.~(\ref{eq:ntxent}) is computed on $\bar{z}_1$ and $\bar{z}_2$ (and their negatives) exactly as in SimCLR. We also evaluated computing the NT-Xent loss independently at each timestep before averaging the resulting per-timestep losses, inspired by TET~\cite{tet}, instead of time-averaging the embeddings first; both orderings yield similar downstream accuracy, so we report only the simpler time-average-then-loss strategy above throughout this paper. Exploiting temporal structure more directly in the contrastive loss is also central to concurrent work: SpikeCLR~\cite{spikeclr} aligns representations across all pairs of timesteps via a Cross Temporal Loss, motivated by the same observation that naively collapsing the temporal dimension before the loss can overlook spike-timing structure. In our case, the input has only $T = 4$ bins, individual events are aggregated into histograms, and the loss is only based on the time-averaged embedding. EventCLR is thus closer to a stateless method than to a stateful one. A stateful model relies on an internal state (here, the membrane potentials) that carries information from one time step to the next, so that its output depends on the order and timing of the input. In our pipeline each bin can almost be treated as a frame: on CIFAR10-DVS, a non-spiking ResNet-18 that processes the bins independently, with no state between time steps, reaches similar accuracy to the SNN (Supplementary Material, Section~S4.3). The LIF neurons do carry a state over the $T = 4$ steps, but on this dataset it does not change the accuracy.

\subsection{Augmentations}

Augmentation design is central to self-supervised learning: the two views must be sufficiently different to avoid trivial matching, while preserving the semantic content required by the downstream task. For event-based data, this choice is less straightforward than for RGB images, since events encode sparse spatio-temporal changes and polarity rather than color channels.

\paragraph{A Generic Pre-/Post-Representation Pipeline.}
We cast event augmentation as a generic four-stage pipeline that decouples \emph{how} events are transformed from \emph{which} grid-based representation is eventually fed to the backbone. A raw \emph{event flux} $\mathcal{E}=\{e_k\}_k$ is first transformed by a \emph{pre-representation augmentation} $\mathcal{A}_{\text{pre}}$, applied directly on the asynchronous event stream (timestamps, coordinates, and polarities) before any spatial quantization. The resulting stream is then mapped to a grid-based tensor by a \emph{representation function} $\rho(\cdot)$ -- the normalized event histogram of Section~\ref{sec:methods} in our experiments, although $\rho(\cdot)$ could equally be a voxel grid or a time surface without altering the rest of the pipeline. Finally, a \emph{post-representation augmentation} $\mathcal{A}_{\text{post}}$ is applied to this grid, where classical image-style transforms become well defined. An augmented view is thus obtained as
\begin{equation}
    x' = \mathcal{A}_{\text{post}}\big(\rho(\mathcal{A}_{\text{pre}}(\mathcal{E}))\big).
\end{equation}
This split is dictated by information availability for two of our three families: transformations that require access to individual event timestamps (\eg, temporal cropping, jitter, or dropping) are only meaningful \emph{before} quantization into a grid, whereas transformations that act on accumulated pixel intensities (\eg, intensity scaling) are only well defined \emph{after} quantization. Spatial transformations (\eg, cropping, flipping, or rolling) are, in contrast, well defined at either stage, since the underlying $(x,y)$ coordinates are available both before and after quantization; we apply rolling to the raw events and cropping and flipping to the histogram. Because $\rho(\cdot)$ is treated as an interchangeable component, a new event representation can be substituted without redesigning the augmentation families, and a new augmentation can be classified as pre- or post-representation according to whether it needs raw event timing or a structured spatial grid, making this pipeline a generic template for event-based data processing beyond the specific families studied here.

In this work, we apply the polarity/intensity family and, from the spatial family, cropping and flipping to the event histogram, and the temporal family and spatial rolling directly to the raw event stream, following the pre-/post-representation split introduced above. Working on histograms for the first group matches SimCLR, where transformations are defined on images, and lets us use standard convolutional backbones without altering the event-to-grid encoding. We group event augmentations into three families:
(i) \emph{spatial} transformations (rolling pre-representation, $\mathcal{A}_{\text{pre}}$; cropping and flipping post-representation, $\mathcal{A}_{\text{post}}$), (ii) \emph{polarity/intensity} transformations (post-representation, $\mathcal{A}_{\text{post}}$), and (iii) \emph{temporal} transformations (pre-representation, $\mathcal{A}_{\text{pre}}$). For each view, $\mathcal{A}_{\text{pre}}$ applies the temporal crop and then rolling, and $\mathcal{A}_{\text{post}}$ applies cropping, flipping, polarity jitter, and polarity averaging in this order; the two views of a sample are augmented independently.

\paragraph{Spatial augmentations.}
We directly reuse classical geometric transforms that preserve object identity while changing viewpoint and spatial support. Rolling is applied to the raw events with a probability of 0.5: each event is shifted cyclically,
\begin{equation}
    x_k' = (x_k + \delta_x) \bmod W_s, \qquad y_k' = (y_k + \delta_y) \bmod H_s,
\end{equation}
where $W_s \times H_s$ is the sensor resolution and the integer shifts are drawn uniformly, $\delta_x \sim \mathcal{U}\{-\lfloor W_s/2 \rfloor, \dots, \lfloor W_s/2 \rfloor\}$ and $\delta_y \sim \mathcal{U}\{-\lfloor H_s/2 \rfloor, \dots, \lfloor H_s/2 \rfloor\}$. On the histogram, we subsequently apply a random resized crop (area scale in $[0.1, 1]$, resized back to $48 \times 48$ with nearest-neighbor interpolation) and a horizontal flip with a probability of 0.5. The same crop and flip are applied to all $T$ bins and both polarity channels. These operations encourage invariance to translation, scale, and left--right symmetry. As illustrated in Figure~\ref{fig:augmentations}(b), rolling translates the object cyclically within the frame, altering the geometric layout while keeping the event content intact.

\paragraph{Polarity and intensity augmentations (colorimetric analogues).}
This family also operates in the post-representation stage $\mathcal{A}_{\text{post}}$. Photometric augmentations used in SimCLR (\eg brightness/contrast/saturation/hue jitter) are not directly meaningful for events. Instead, we interpret their role as inducing controlled changes in the distribution of pixel intensities while preserving structure. For event histograms, we implement this effect with a polarity jitter applied with a probability of 0.8. It combines a global scaling (brightness-like) and a scaling around the mean of the two polarity channels (contrast-like), applied in random order:
\begin{equation}
    \bar{\mathcal{H}}' = b\,\bar{\mathcal{H}}, \qquad \bar{\mathcal{H}}'[n, c, y, x] = a\left(\bar{\mathcal{H}}[n, c, y, x] - \mu[n, y, x]\right) + \mu[n, y, x],
\end{equation}
where $\mu[n, y, x] = \frac{1}{2}\sum_{c} \bar{\mathcal{H}}[n, c, y, x]$, and the factors $a, b \sim \mathcal{U}[0.5, 1.5]$ are drawn once per view. No clipping is applied. Finally, with a probability of 0.2, polarity averaging replaces both channels by their mean $\mu$, which removes the polarity information from the view. These transforms encourage robustness to sensor gain and threshold changes. Figure~\ref{fig:augmentations}(c) shows how brightness scaling visibly changes the intensity distribution across the two polarity channels without modifying the spatial structure.

\paragraph{Temporal augmentations.}
This family operates in the pre-representation stage $\mathcal{A}_{\text{pre}}$: while our main pipeline operates on event histograms, temporal variability is introduced upstream of histogram construction, directly on the raw event flux, by sampling a random temporal window (temporal crop) from the event stream. This encourages invariance to event timing and motion phase while keeping the underlying content consistent. In practice, for each view we draw a crop ratio $r \sim \mathcal{U}[0.1, 1]$ and a start time $t_s \sim \mathcal{U}[t_0, t_{\text{end}} - r\,(t_{\text{end}} - t_0)]$, keep the events with $t_s \le t_k < t_s + r\,(t_{\text{end}} - t_0)$, and convert the resulting window to a histogram via $\rho(\cdot)$, so that the $T$ bins of Eq.~(\ref{eq:histogram}) cover the cropped window. As shown in Figure~\ref{fig:augmentations}(d), cropping a different time window captures a different phase of the camera motion, producing a histogram with a noticeably different event distribution yet the same underlying semantic content.

Once converted into histograms, event data can be processed like images. The main difference with RGB images is the visual content, mostly edges and motion rather than color and texture, so the SimCLR recipe can be transferred if the augmentations are adapted to this content. Overall, our augmentation pipeline aims to mimic the function of SimCLR augmentations, learning invariances to geometry, intensity statistics, and temporal sampling, while respecting the sparse, polarity-based nature of event data. Figure~\ref{fig:augmentations} provides a visual comparison of these three augmentation families. Because these families are organized around the generic pre-/post-representation split introduced above rather than around the specific choice of $\rho(\cdot)$, the same pipeline applies unchanged if the event histogram is replaced by another grid-based representation, or if additional pre- or post-representation transformations are introduced.

\subsection{Evaluation Protocol}
\label{sec:protocol}

We evaluate EventCLR on five event-based datasets, using a two-phase protocol of self-supervised pretraining followed by downstream evaluation to assess data efficiency and representation quality.

\paragraph{Experimental Protocol.} In the first phase, we pretrain a SEW-ResNet-18 backbone without using any labels, following the SimCLR contrastive approach. We employ the three augmentation families described above: spatial, polarity/intensity, and temporal augmentations. For the event representation, we primarily use normalized event histograms, but we also evaluated non-normalized histograms and Voxel Grid \cite{voxelgrid} representations, finding that the overall behavior and conclusions remain unchanged. Following standard SimCLR practice, the projection head $g(\cdot)$ is discarded after pretraining; only the encoder $f(\cdot)$ is retained for downstream evaluation. Pretraining uses only the training split of each dataset: validation and test recordings are never seen during pretraining. We pretrain for 500 epochs (1,000 for DVS-Gesture) with a batch size of $N = 256$, a temperature $\tau = 0.2$, and AdamW (learning rate $10^{-3}$, weight decay $10^{-4}$) with a cosine schedule down to 1\% of the initial learning rate. Our SNN models are implemented using SpikingJelly~\cite{spikingjelly}, and event-specific data augmentations are applied via the Tonic library~\cite{tonic}. Detailed results for alternative event representations, along with all implementation details, are provided in the Supplementary Material. In the second phase, we evaluate the quality of the learned representations using two standard protocols:
\begin{enumerate}
    \item \textbf{Linear Probing (LP):} The backbone weights are frozen, and only a single linear classifier is trained on top of the representations. This protocol directly probes the \emph{linear separability} of the learned feature space: if classes are linearly separable without any further adaptation of the backbone, it demonstrates that the encoder has captured semantically meaningful structure during pretraining alone. Training only a linear head is also cheaper than retraining the full SNN backbone, which could be relevant for on-device adaptation, although we do not evaluate this aspect here. In our implementation, BatchNorm statistics in the backbone are still updated during linear probing, and the same augmentations used for fine-tuning are applied.
    \item \textbf{Fine-tuning (FT):} The entire network (backbone and linear classifier) is fine-tuned end-to-end. This evaluates the transferability of the pretrained initialization for downstream supervised tasks, allowing the model to adapt its representations to the target distribution.
\end{enumerate}
As a baseline, we compare against fully supervised learning trained from scratch (Supervised). All downstream models (LP, FT, and Supervised) are trained for 150 epochs with a batch size of 64 and AdamW (learning rate $10^{-3}$, weight decay $10^{-4}$, cosine schedule), and use the same downstream augmentations: a random temporal crop keeping a fraction $r \sim \mathcal{U}[0.5, 1]$ of the recording, rolling by up to 20\% of the sensor size with a probability of 0.5, CutMix with a probability of 0.5, and, with a probability of 0.1, a light post-histogram stage (random resized crop, horizontal flip, and polarity jitter; Supplementary Material, Section~S2). These augmentations are lighter than the pretraining ones. Training the supervised baseline with the full EventCLR augmentations instead lowers its accuracy on both CIFAR10-DVS and DVS-Gesture (Supplementary Material, Section~S4.2), so the gains of EventCLR do not come from its augmentations alone. To assess data efficiency, we perform evaluations under various labeled data regimes, ranging from few-shot scenarios ($k \in \{1, 10, 20, 50\}$ samples per class) to fractions of the labeled training set. For all experiments, we report the mean and standard deviation across independent random label splits, using 10 splits to increase robustness over the random label selection when computationally feasible, and 3 splits otherwise to limit computational cost. No separate labeled validation set is used. Each dataset has a single held-out split: the official test set for N-MNIST, DVS-Gesture and N-Cars, and the 10\% stratified split of Section~\ref{sec:datasets} for CIFAR10-DVS and N-Caltech101. The same split serves to select the epoch and to report accuracy: for every method, including the supervised baseline, we report the highest accuracy reached on it over the training epochs, since models trained on a few labeled samples tend to overfit in later epochs. The labeled training subset is the only labeled data used for fitting, and pretraining never sees the held-out split. Because the epoch is selected on the evaluation data, absolute accuracies are optimistic; the same rule is applied to all methods, so comparisons between them are on equal footing.

\section{Experiments}

We report results on all five event-based datasets under the evaluation protocol described in Section~\ref{sec:protocol}.

\subsection{Overview of Main Results}
\label{sec:overview}

\setlength{\tabcolsep}{10pt}
\begin{table}[t]
    \centering
    \caption{Overview of few-shot fine-tuning accuracy (\%, mean $\pm$ std over 3 random subsets, 10 for N-MNIST) across all five benchmarks. \textbf{EventCLR} denotes self-supervised pretraining followed by fine-tuning; \textbf{Supervised only} denotes training from scratch. $\Delta$ is the absolute gain of EventCLR over Supervised only; the value in parentheses is the relative error reduction (RER), i.e.\ the fraction of the Supervised-only baseline's remaining error that EventCLR closes, which accounts for how close the baseline already is to the 100\% ceiling.}
    \label{tab:main_results}
    \begin{threeparttable}
    \begin{tabular}{ll|ccc}
        \toprule
        Dataset & Method & 1-shot & 10-shot & 50-shot \\
        \midrule
        \multirow{3}{*}{CIFAR10-DVS}
            & EventCLR   & \textbf{21.2} $(\pm 3.6)$ & \textbf{38.5} $(\pm 0.1)$ & \textbf{51.7} $(\pm 0.6)$ \\
            & Supervised only & 13.5 $(\pm 1.2)$ & 29.0 $(\pm 2.2)$ & 44.8 $(\pm 0.5)$ \\
            & $\Delta$   & \gain{7.7}~\relgain{8.9} & \gain{9.5}~\relgain{13.4} & \gain{6.9}~\relgain{12.5} \\
        \midrule
        \multirow{3}{*}{N-Caltech101}
            & EventCLR   & \textbf{21.3} $(\pm 3.8)$ & \textbf{51.6} $(\pm 0.8)$ & \textbf{73.3} $(\pm 0.6)$ \\
            & Supervised only & 7.9 $(\pm 1.0)$ & 45.6 $(\pm 0.4)$ & 70.4 $(\pm 0.7)$ \\
            & $\Delta$   & \gain{13.4}~\relgain{14.5} & \gain{6.0}~\relgain{11.0} & \gain{2.9}~\relgain{9.8} \\
        \midrule
        \multirow{3}{*}{DVS-Gesture}
            & EventCLR   & \textbf{37.8} $(\pm 1.4)$ & \textbf{68.3} $(\pm 2.1)$ & \textbf{85.4} $(\pm 2.1)$ \\
            & Supervised only & 22.7 $(\pm 2.4)$ & 61.7 $(\pm 2.3)$ & 83.1 $(\pm 1.3)$ \\
            & $\Delta$   & \gain{15.1}~\relgain{19.5} & \gain{6.6}~\relgain{17.2} & \gain{2.3}~\relgain{13.6} \\
        \midrule
        \multirow{3}{*}{N-Cars}
            & EventCLR   & \textbf{60.9} $(\pm 4.3)$ & \textbf{79.3} $(\pm 2.4)$ & \textbf{88.1} $(\pm 1.2)$ \\
            & Supervised only & 54.8 $(\pm 3.7)$ & 60.5 $(\pm 0.3)$ & 79.1 $(\pm 0.9)$ \\
            & $\Delta$   & \gain{6.1}~\relgain{13.5} & \gain{18.8}~\relgain{47.6} & \gain{9.0}~\relgain{43.1} \\
        \midrule
        \multirow{3}{*}{N-MNIST}
            & EventCLR   & \textbf{60.5} $(\pm 6.5)$ & \textbf{94.1} $(\pm 0.7)$ & \textbf{97.6} $(\pm 0.2)$ \\
            & Supervised only & 21.2 $(\pm 3.6)$ & 84.3 $(\pm 1.6)$ & 95.7 $(\pm 0.4)$ \\
            & $\Delta$   & \gain{39.3}~\relgain{49.9} & \gain{9.8}~\relgain{62.4} & \gain{1.9}~\relgain{44.2} \\
        \bottomrule
    \end{tabular}
    \begin{tablenotes}
        \footnotesize
        \item[] RER (relative error reduction) $= \Delta / (100 - \text{Supervised only})$: the share of the supervised-only baseline's remaining error eliminated by EventCLR pretraining.
    \end{tablenotes}
    \end{threeparttable}
\end{table}

Table~\ref{tab:main_results} previews our main results, consolidating fine-tuning accuracy across all five benchmarks, CIFAR10-DVS, N-Caltech101, DVS-Gesture, N-Cars, and N-MNIST, in the 1-shot, 10-shot, and 50-shot regimes. EventCLR pretraining outperforms training from scratch in every single configuration, and for most datasets the gain is largest where labels are scarcest: +39.3 points on N-MNIST, +15.1 on DVS-Gesture, and +13.4 on N-Caltech101 in the 1-shot setting. N-Cars departs from this pattern: its largest absolute gain occurs at 10-shot (+18.8 points, 47.6\% RER) rather than at 1-shot (+6.1 points), and a milder version of the same effect is visible on CIFAR10-DVS, whose gain also peaks at 10-shot (+9.5) rather than 1-shot (+7.7). As more labeled data becomes available the gap eventually narrows, but it never closes by 50-shot, confirming that self-supervised pretraining remains valuable beyond the most extreme label-scarce regime this work targets. The remainder of this section examines each of these results in detail, together with the augmentation choices, backbone architectures, and pretraining conditions that drive them. The Linear Probing counterpart of Table~\ref{tab:main_results}, together with the same $\Delta$/RER analysis, is provided in the Supplementary Material.

\subsection{Results on Static-Image-Derived Datasets}

We first evaluate performance on CIFAR10-DVS, where the model is both pretrained and evaluated on the same dataset. This in-domain setting allows us to assess the effectiveness of our self-supervised approach when unlabeled and labeled data come from the same distribution.

\begin{figure*}[t]
    \centering
    \subfloat[]{
    \includegraphics[width=0.40\linewidth]{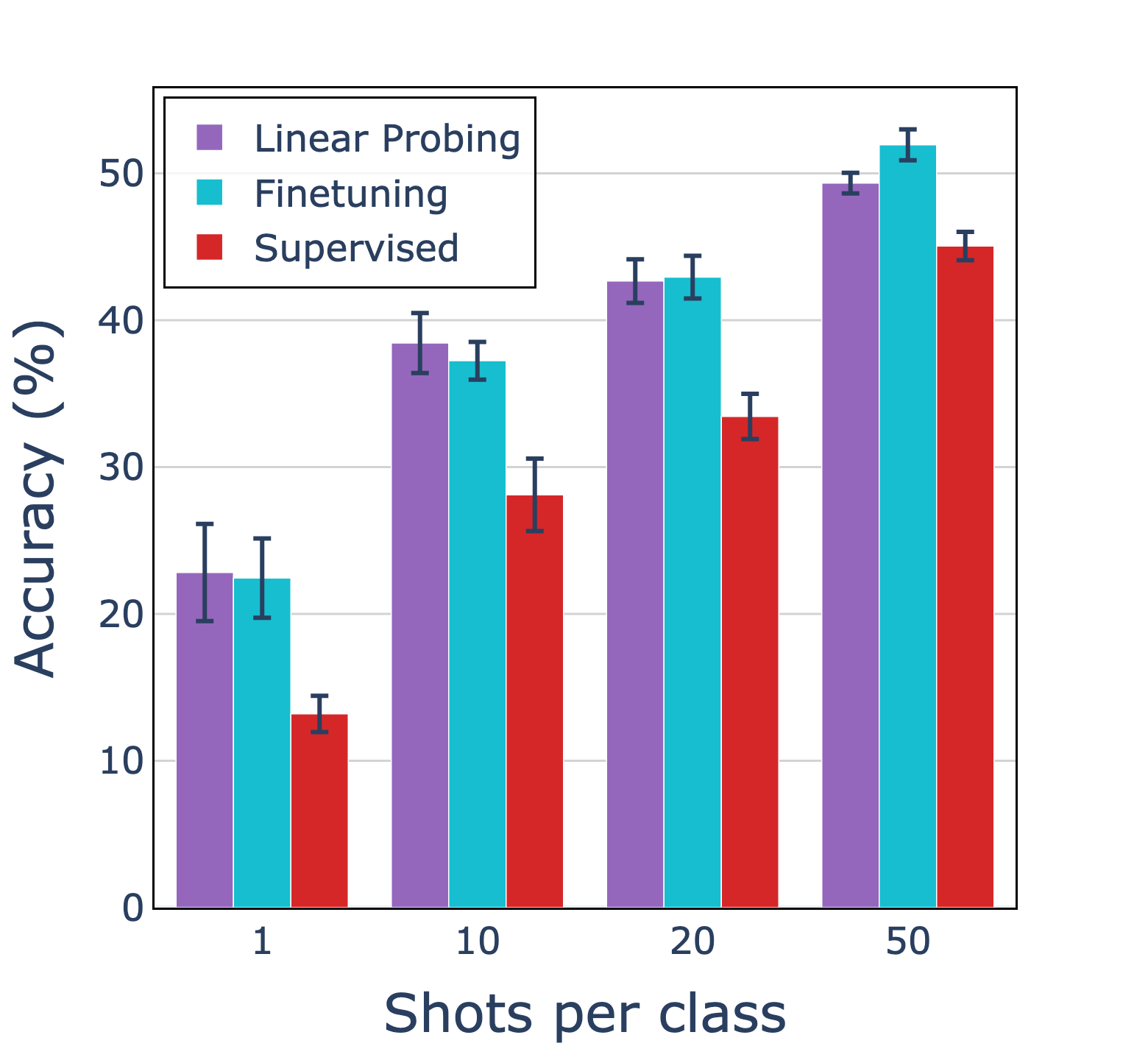}
        \label{fig:cifar10dvs_fewshot}
    }
    \subfloat[]{
    \includegraphics[width=0.40\linewidth]{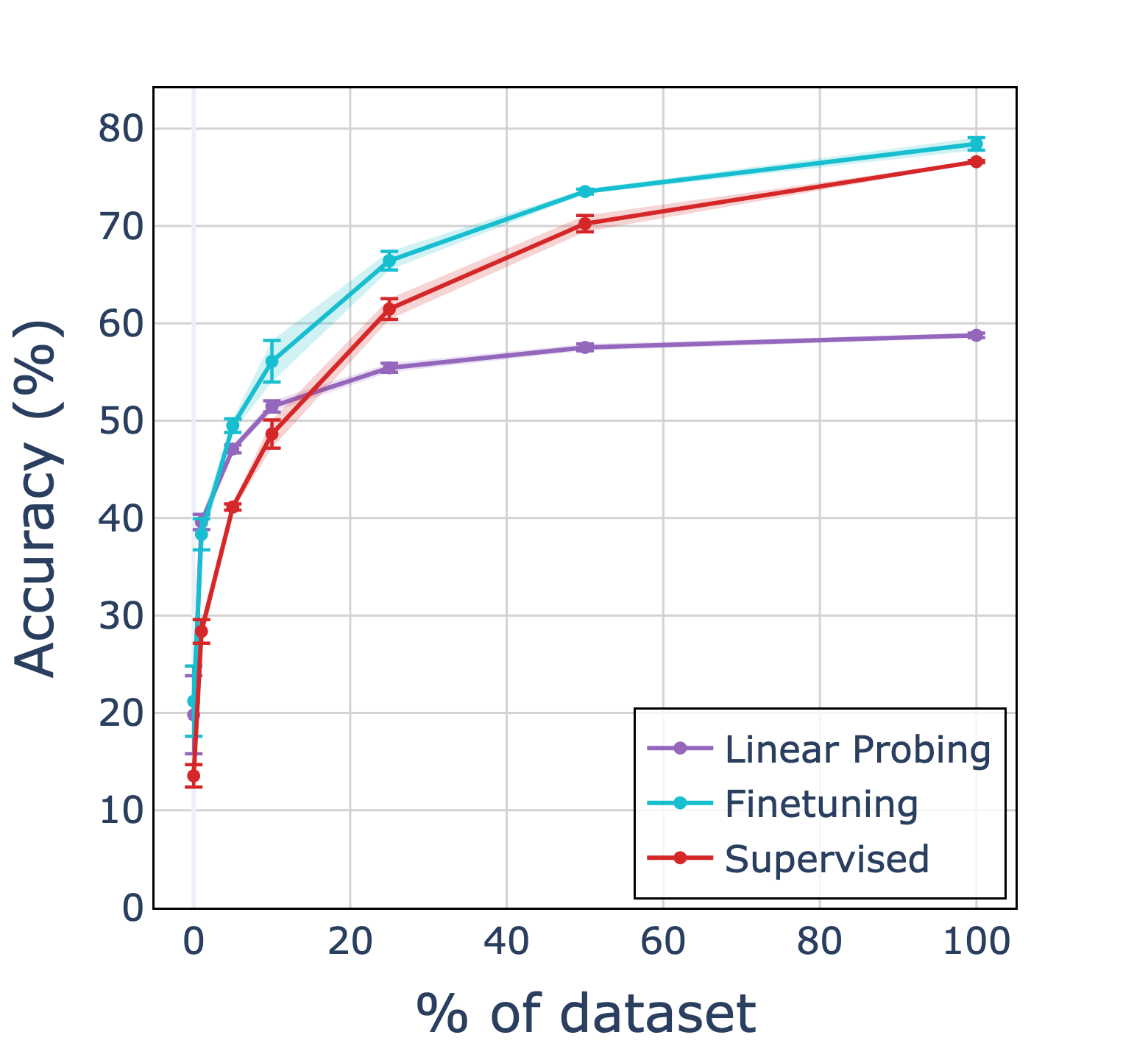}
        \label{fig:cifar10dvs_evolution}
    }\\
    \subfloat[]{
        \includegraphics[width=0.40\linewidth]{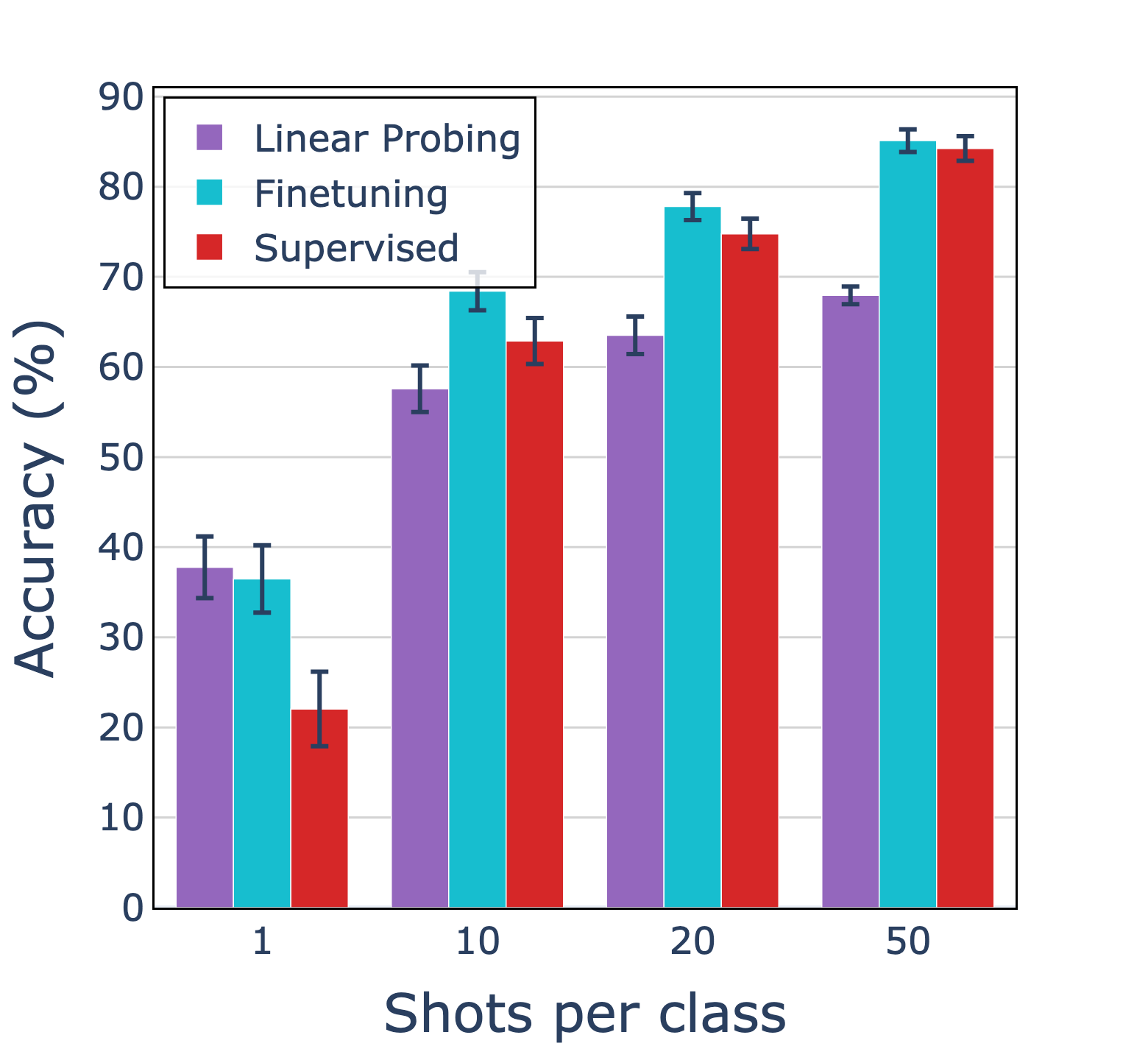}
        \label{fig:DVS-Gesture_fewshot}
    }
    \subfloat[]{
        \includegraphics[width=0.40\linewidth]{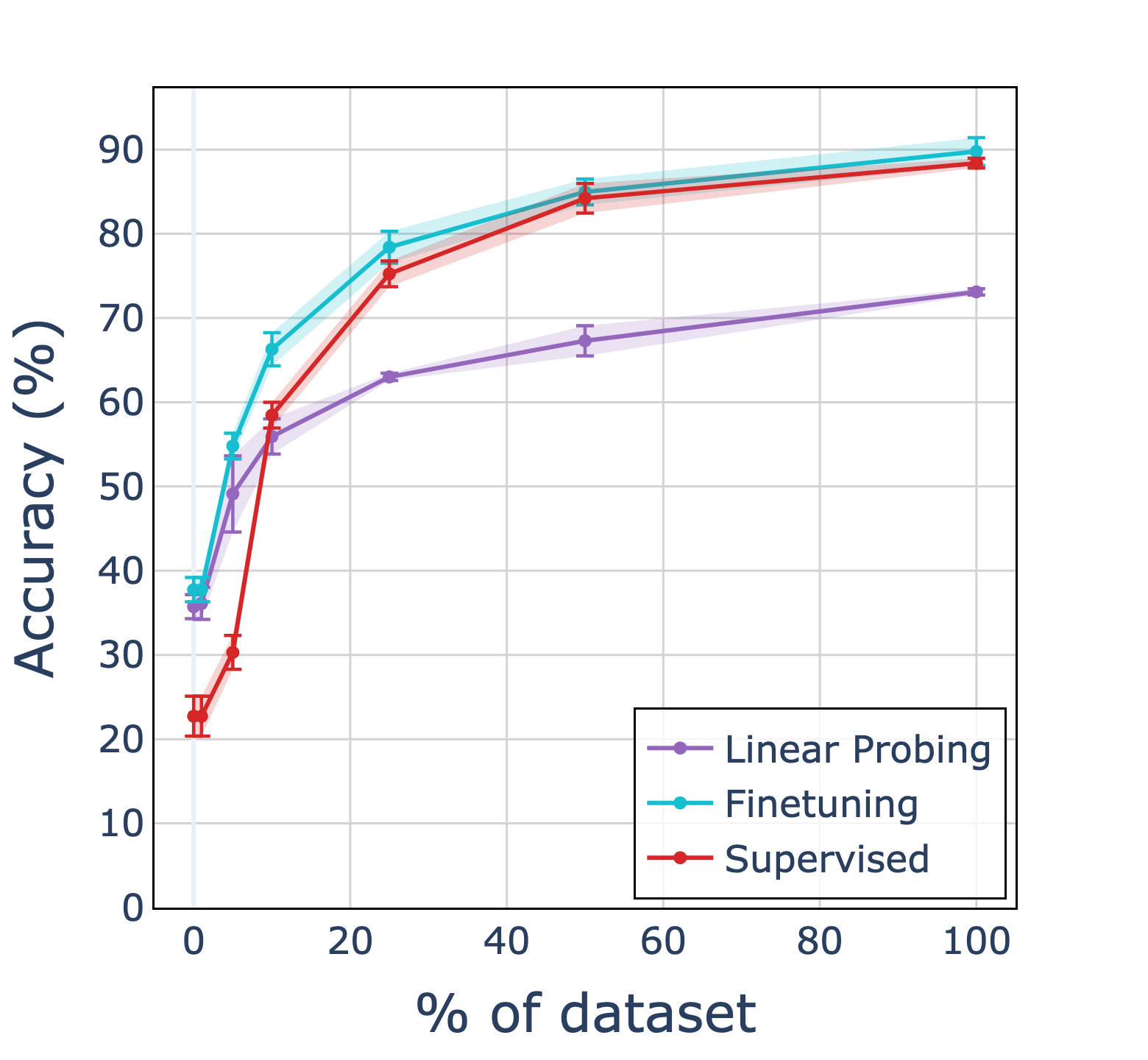}
        \label{fig:DVS-Gesture_evolution}
    }
    \caption{Classification accuracy on CIFAR10-DVS (a, b) and DVS-Gesture (c, d). (a, c)~Few-shot comparison across methods (mean and std over 10 random subsets). (b, d)~Accuracy as a function of the percentage of labeled training data (mean and std over 3 random subsets).}
    \label{fig:cifar10dvs_results}
    \label{fig:DVS-Gesture_results}
\end{figure*}

Figure~\ref{fig:cifar10dvs_results} presents the quantitative results. In the few-shot regime (Figure~\ref{fig:cifar10dvs_fewshot}), our self-supervised method consistently outperforms the supervised baseline for both Linear Probing and Fine-Tuning protocols. This demonstrates that contrastive pretraining yields robust features that generalize well even with limited supervision. The performance gap is particularly pronounced in the most extreme low-data scenarios, where self-supervised pretraining provides the encoder with rich prior knowledge about event-based spatio-temporal patterns.

As the amount of labeled data increases (Figure~\ref{fig:cifar10dvs_evolution}), fine-tuning remains superior to supervised learning across all regimes, validating the benefits of self-supervised initialization. The pretrained representations provide a better starting point, enabling faster convergence and higher final accuracy. Linear Probing is competitive in low-data regimes, demonstrating linear separability of the features, but saturates earlier than FT or supervised learning when fine-grained adaptation is needed.

We observe the same qualitative trends on N-Caltech101 for fine-tuning, while linear probing falls below the supervised baseline from 10-shot on; the corresponding results are provided in the Supplementary Material.

\begin{figure}[t]
    \centering
    \includegraphics[width=\linewidth]{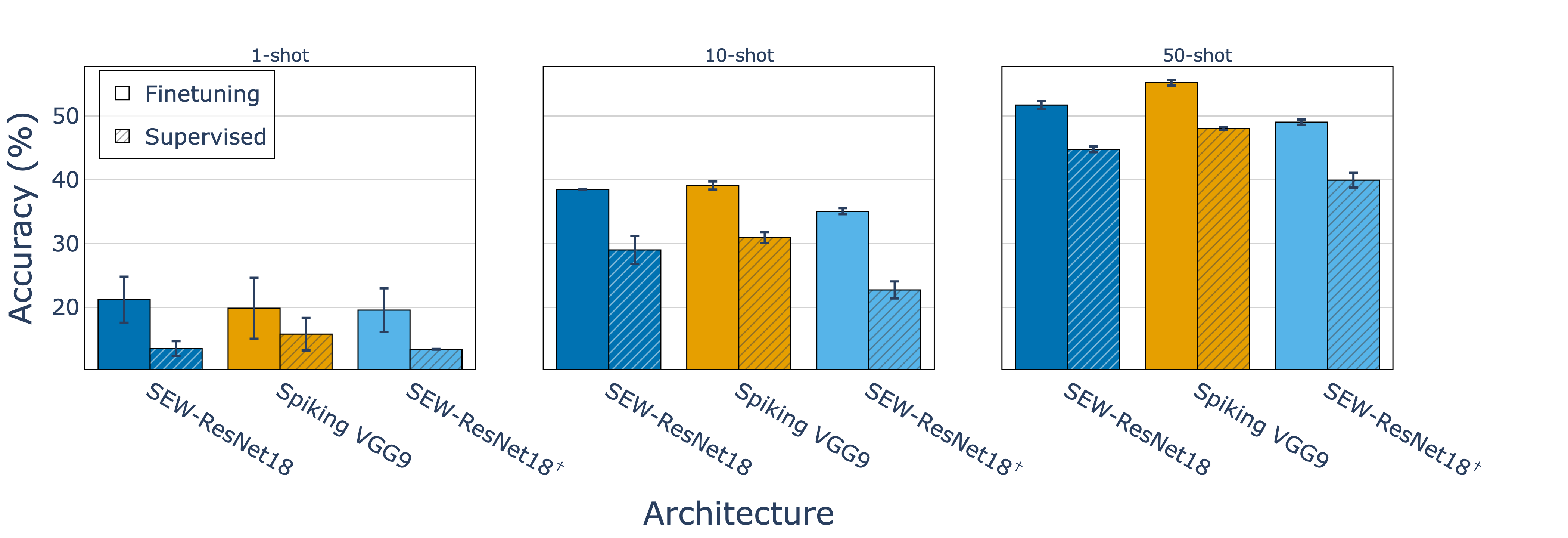}
    \caption{Few-shot accuracy on CIFAR10-DVS across different SNN backbones (averaged across 3 random subsets). ${}^{\dag}$ denote separable convolution.}
    \label{fig:archi_results}
\end{figure}

\paragraph{Impact of Backbone Architecture.} To assess the generality of our approach across different network architectures, we evaluate three SNN backbones with varying capacities and computational characteristics. Figure~\ref{fig:archi_results} reports the few-shot accuracy for each configuration. Results demonstrate that the benefits of self-supervised pretraining generalize across different architectures. All three backbones exhibit the same qualitative behavior: fine-tuned models consistently outperform their supervised counterparts across all few-shot regimes. The SEW-ResNet-18 reaches 21.2\%, 38.5\%, and 51.7\% (Table~\ref{tab:ablation_aug}). The SEW-ResNet-18 with separable convolutions achieves competitive accuracy (19.6\%, 35.1\%, 49.0\%) while reducing the parameter count by 8$\times$ (from 11.2M to 1.4M), making it particularly attractive for resource-constrained or on-device deployments. The Spiking VGG9 (9.2M parameters) architecture yields the strongest overall performance, reaching 55.2\% in the 50-shot scenario and achieving 81.0\% accuracy when fine-tuned on the full CIFAR10-DVS dataset. On CIFAR10-DVS, a standard non-spiking ResNet-18 trained with the same pipeline reaches a similar accuracy and also benefits from pretraining (Supplementary Material, Section~S4.3), so the gains of EventCLR are not specific to SNNs. These results confirm that our self-supervised approach is architecture-agnostic and can be effectively applied to diverse SNN designs.

\subsection{Ablation Study}
\label{sec:ablattion}

\paragraph{Augmentations} A critical component of SimCLR is the composition of data augmentations. We conduct an ablation study to identify which transformations are most important for learning invariant event representations. We evaluate performance in 1-shot, 10-shot, and 50-shot scenarios.

Results in Table~\ref{tab:ablation_aug} reveal the distinct contributions of different augmentation families. When used in isolation, spatial augmentations are consistently the weakest performer (16.0\%, 29.3\%, 44.4\%), yielding only marginal, and at times negative, gains over the supervised baseline (+2.5\%, +0.3\%, and -0.4\% across the three settings). This suggests that spatial invariances, while important, are less critical for event-based representations than their temporal and polarity counterparts. Temporal augmentations achieve the best single-family performance in the 10-shot and 50-shot regimes (31.9\%, 48.5\%), improving over the supervised baseline by +2.9\% and +3.7\%, respectively, which underscores the importance of learning temporal invariances for processing the inherently dynamic nature of event streams. Polarity augmentations lead in the most extreme 1-shot setting (19.2\%, +5.7\% over supervised) and remain competitive elsewhere (30.2\%, 46.5\%), demonstrating that robustness to polarity variations is also beneficial for representation quality.

\setlength{\tabcolsep}{10pt}
\begin{table}[t]
    \centering
    \caption{Ablation study of augmentation families on CIFAR10-DVS. Few-shot fine-tuning accuracy (\%) reported as mean $\pm$ std over 3 random subsets.}
    \label{tab:ablation_aug}
    \begin{threeparttable}
    \begin{tabular}{l|ccc}
        \toprule
        Augmentation Set & 1-shot & 10-shot & 50-shot \\
        \midrule
        EventDrop-style set$^\dag$ \tcite{eventdrop} & 18.1 $(\pm 1.3)$ & 31.2 $(\pm 1.8)$ & 47.6 $(\pm 0.9)$ \\
        NDA-style set$^\dag$ \tcite{nda} & 15.9 $(\pm 2.7)$ & 29.8 $(\pm 1.6)$ & 44.9 $(\pm 1.3)$ \\
        \midrule
        Spatial only & 16.0 $(\pm 2.1)$ & 29.3 $(\pm 1.7)$ & 44.4 $(\pm 1.2)$ \\
        Temporal only & 17.3 $(\pm 1.6)$ & 31.9 $(\pm 0.8)$ & 48.5 $(\pm 1.0)$ \\
        Polarity only & 19.2 $(\pm 1.7)$ & 30.2 $(\pm 2.0)$ & 46.5 $(\pm 0.8)$ \\
        Spatial + Temporal & \underline{20.2} $(\pm 3.9)$ & \underline{36.9} $(\pm 0.5)$ & \underline{50.7} $(\pm 1.1)$ \\
        Spatial + Polarity & 19.2 $(\pm 3.6)$ & 34.0 $(\pm 1.1)$ & 48.7 $(\pm 0.5)$ \\
        Temporal + Polarity & 17.9 $(\pm 2.3)$ & 32.5 $(\pm 0.7)$ & 49.0 $(\pm 0.4)$ \\
        \textbf{Full Pipeline} & \textbf{21.2} $(\pm 3.6)$ & \textbf{38.5} $(\pm 0.1)$ & \textbf{51.7} $(\pm 0.6)$ \\
        \midrule
        Supervised & 13.5 $(\pm 1.2)$ & 29.0 $(\pm 2.2)$ & 44.8 $(\pm 0.5)$ \\
        \bottomrule
    \end{tabular}
    \begin{tablenotes}
        \footnotesize
        \item[$\dag$] Our implementation of the augmentation policy only, used as the contrastive augmentation during pretraining; these rows are not the full EventDrop or NDA methods.
    \end{tablenotes}
    \end{threeparttable}
\end{table}

Combining augmentation families leads to synergistic improvements. The spatial and temporal combination is the second-best configuration across all three settings (20.2\%, 36.9\%, 50.7\%), confirming that these two families complement each other effectively. The full augmentation pipeline, incorporating all three families, achieves the overall best performance across all few-shot settings (21.2\%, 38.5\%, and 51.7\%), with consistent gains of +7.7\%, +9.5\%, and +6.9\% over the supervised baseline. The ranking of the combined sets is clear at 10-shot; at 1-shot and 50-shot, the gap between the full pipeline and Spatial + Temporal is within one standard deviation. This validates our design choice to leverage the unique characteristics of event data through specialized augmentations across spatial, temporal, and polarity dimensions.

We also pretrain with the augmentation sets of EventDrop~\cite{eventdrop} and NDA~\cite{nda}, implemented by us and used as contrastive augmentations; this compares augmentation policies, not the full methods. Both improve over the supervised baseline, but neither matches our full pipeline in the few-shot regime, suggesting that augmentations optimised for classification do not straightforwardly transfer to the contrastive setting. This gap is consistent with the ablation: NDA is essentially a spatial strategy and EventDrop acts as a temporal one, so their individual performance mirrors that of the corresponding single-family variants in our study.

\paragraph{SimCLR hyperparameters} Varying the temperature ($\tau \in \{0.1, 0.2, 0.5\}$) and the pretraining batch size ($N$ from 8 to 256) changes few-shot fine-tuning accuracy on CIFAR10-DVS by at most 3 points, which is close to the variation across random subsets. In particular, reducing the batch size to 8 does not degrade the results. These results are reported in the Supplementary Material (Section~S4.1).

\subsection{Results on Dynamic Event-Stream Datasets}

To further validate generalizability, we evaluate on DVS-Gesture, which contains naturally recorded event streams. Figures~\ref{fig:DVS-Gesture_fewshot} and~\ref{fig:DVS-Gesture_evolution} report the results. In the few-shot regime (Figure~\ref{fig:DVS-Gesture_fewshot}), fine-tuning from EventCLR improves over training from scratch at every label budget. Linear probing only helps at 1-shot: from 10-shot on, it is below the supervised baseline, so on this dataset the frozen features learned from a small unlabeled set are not sufficient on their own. As more labels become available (Figure~\ref{fig:DVS-Gesture_evolution}), the gap with the supervised baseline progressively shrinks, becoming marginal beyond ${\sim}50\%$ of the training set, which we attribute to the small dataset size (1,077 training sequences). These results confirm that EventCLR is effective for both re-display and dynamic event-stream datasets. We additionally evaluate it on N-Cars, a second naturally dynamic dataset recorded with a different sensor (ATIS) and a different motion source (ego-vehicle rather than subject-driven gestures); the corresponding few-shot results are provided in the Supplementary Material.

\subsection{Transfer Learning}

To assess cross-dataset transfer, we evaluate models pretrained on one event dataset and adapted to a different target dataset. We consider three target domains (CIFAR10-DVS, DVS-Gesture, and N-Caltech101) and for each we compare in-domain pretraining, cross-domain pretraining, and training from scratch. This protocol tests whether the learned spatio-temporal features capture dataset-agnostic structure rather than overfitting to a single acquisition protocol. This setting is practically relevant when unlabeled data are available in one domain but labels are scarce in the deployment domain.

In our transfer experiments, we primarily focus on \emph{static-to-dynamic} transfer: models pretrained on a static-image-derived dataset (CIFAR10-DVS or N-Caltech101) are adapted to a dynamic dataset (DVS-Gesture), but not the reverse. We complement this with a \emph{dynamic-to-dynamic} adaptation, pretraining on N-Cars~\cite{timesurface2} and adapting to DVS-Gesture, to test whether transfer benefits also hold between two naturally dynamic datasets recorded with different sensors (ATIS \vs DVS) and with a different motion source (ego-vehicle \vs subject-driven gestures), rather than only from static-image-derived to dynamic data. This direction is the most practically relevant, as large collections of re-display recordings are far easier to obtain than labeled real-motion sequences. Note that CIFAR10-DVS and N-Caltech101 were recorded with different hardware and motion patterns, so the transfer between them is non-trivial.

We also pretrain EventCLR on N-ImageNet~\cite{nimagenet} for 10 epochs and fine-tune it on CIFAR10-DVS, DVS-Gesture, N-Caltech101, and N-Cars; these results, obtained on different hardware, are reported in the Supplementary Material (Section~S4.4). With only 10 epochs of pretraining, N-ImageNet improves over training from scratch on the four targets; it is on par with in-domain pretraining on DVS-Gesture and N-Caltech101 at every label budget, but remains below in-domain pretraining on CIFAR10-DVS at 10 and 50 shots and on N-Cars at every label budget, where MEM~\cite{mem} reports the same trend. Event Camera Data Pre-training~\cite{eventpretraining} reports results in a similar setting, but it uses a non-spiking backbone and paired RGB images during pretraining, so we do not compare accuracies directly. For context, full-label accuracies of EventCLR and values reported by other event-based methods on N-Caltech101 and N-Cars are given in the Supplementary Material (Section~S4.5).

\begin{table}[t]
    \centering
    \caption{Cross-dataset transfer learning results. Few-shot fine-tuning accuracy (\%) reported as mean $\pm$ std over 3 random subsets. Models pretrained on different event datasets are fine-tuned on distinct target datasets and evaluated in few-shot scenarios.}
    \label{tab:transfer_learning}
    \begin{tabular}{c c|ccc}
        \toprule
        \multicolumn{2}{c}{Dataset} & 1-shot & 10-shot & 50-shot \\
        Pretrain & Target & & & \\
        \midrule
        CIFAR10-DVS & CIFAR10-DVS & 21.2 $(\pm 3.6)$ & 38.5 $(\pm 0.1)$ & 51.7 $(\pm 0.6)$ \\
        N-Caltech101 & CIFAR10-DVS & 18.9 $(\pm 2.6)$ & 32.0 $(\pm 0.7)$ & 48.0 $(\pm 0.9)$ \\
         - & CIFAR10-DVS & 13.5 $(\pm 1.2)$ & 29.0 $(\pm 2.2)$ & 44.8 $(\pm 0.5)$ \\
        \midrule
        DVS-Gesture & DVS-Gesture & 37.8 $(\pm 1.4)$ & 68.3 $(\pm 2.1)$ & 85.4 $(\pm 2.1)$ \\
        CIFAR10-DVS & DVS-Gesture & 35.0 $(\pm 1.3)$ & 69.7 $(\pm 3.6)$ & 85.7 $(\pm 1.2)$ \\
        N-Cars & DVS-Gesture & 33.8 $(\pm 2.7)$ & 68.6 $(\pm 3.1)$ & 85.6 $(\pm 2.4)$ \\
         - & DVS-Gesture & 22.7 $(\pm 2.4)$ & 61.7 $(\pm 2.3)$ & 83.1 $(\pm 1.3)$ \\
        \midrule
        N-Caltech101 & N-Caltech101 & 21.3 $(\pm 3.8)$ & 51.6 $(\pm 0.8)$ & 73.3 $(\pm 0.6)$ \\
        CIFAR10-DVS & N-Caltech101 & 17.0 $(\pm 2.2)$ & 50.8 $(\pm 0.4)$ & 73.0 $(\pm 0.3)$ \\
         - & N-Caltech101 & 7.9 $(\pm 1.0)$ & 45.6 $(\pm 0.4)$ & 70.4 $(\pm 0.7)$ \\
        \bottomrule
    \end{tabular}
\end{table}

Results in Table~\ref{tab:transfer_learning} show that EventCLR representations transfer across datasets. When pretraining on N-Caltech101 and adapting to CIFAR10-DVS, we observe consistent gains over training from scratch in the few-shot regime, indicating that the encoder captures reusable spatio-temporal primitives. Conversely, pretraining on CIFAR10-DVS also improves few-shot performance on N-Caltech101 compared to the supervised-from-scratch baseline, despite the larger number of classes and different recording statistics. For the dynamic-to-dynamic direction, pretraining on N-Cars and transferring to DVS-Gesture also outperforms training from scratch at every label budget (33.8 \vs 22.7 at 1-shot, 68.6 \vs 61.7 at 10-shot, 85.6 \vs 83.1 at 50-shot), and is essentially on par with in-domain DVS-Gesture pretraining (37.8, 68.3, 85.4) and with CIFAR10-DVS$\to$DVS-Gesture transfer (35.0, 69.7, 85.7) at 10-shot and 50-shot, though it trails both at 1-shot. This confirms that transfer benefits are not specific to the static-to-dynamic direction: they also hold between two naturally dynamic datasets recorded with different sensors (ATIS \vs DVS) and with different motion sources (ego-vehicle \vs subject-driven gestures).

Across CIFAR10-DVS and N-Caltech101 datasets, in-domain pretraining remains better. Cross-dataset transfer is further constrained by domain-specific event statistics and acquisition artifacts. Nevertheless, the cross-dataset improvements confirm that self-supervised pretraining provides a strong generic initialization when labeled data in the target domain are scarce. DVS-Gesture behaves differently: both cross-domain sources, CIFAR10-DVS and N-Cars, are on par with in-domain DVS-Gesture pretraining at 10-shot and 50-shot (differences within one standard deviation), despite differing substantially in acquisition protocol and, for CIFAR10-DVS, in motion source, while in-domain pretraining remains best at 1-shot. This might be due to the difference in the amount of pretraining data, since both CIFAR10-DVS (${\sim}$10K samples) and N-Cars (15,422 training samples) offer a substantially larger unlabeled pretraining corpus than DVS-Gesture's 1,077 training sequences. A related dependence on the pretraining set is reported for MEM~\cite{mem}: pretraining on the much larger N-ImageNet improves N-Caltech101 but not N-Cars, where in-domain pretraining remains better. More unlabeled data therefore helps mainly when the in-domain set is small or when the statistics of the two datasets are close. This observation motivates studying how increasing the amount of unlabeled pretraining data affects the learned representations, which we investigate next.

\subsection{Effect of Unlabeled Data Quantity}

Finally, we analyze how the size of the pretraining dataset affects representation quality. Using N-MNIST, we compare pretraining on the full dataset (60K samples) versus a reduced subset (10K samples), then evaluate the learned representations in few-shot classification scenarios.

\begin{table}[h]
    \centering
    \caption{Impact of pretraining dataset size on N-MNIST. Few-shot fine-tuning accuracy (\%) reported as mean $\pm$ std over 10 random subsets. Models pretrained with different amounts of unlabeled data are compared against supervised baselines.}
    \label{tab:data_quantity}
    \begin{tabular}{l|ccc}
        \toprule
        Amount of pretrained data & 1-shot & 10-shot & 50-shot \\
        \midrule
        10K & 56.9 $(\pm 6.4)$ & 92.1 $(\pm 0.9)$ & 96.8 $(\pm 0.3)$ \\
        60K & 60.5 $(\pm 6.5)$ & 94.1 $(\pm 0.7)$ & 97.6 $(\pm 0.2)$ \\
        \midrule
        Supervised & 21.2 $(\pm 3.6)$  & 84.3 $(\pm 1.6)$ & 95.7 $(\pm 0.4)$ \\
        \bottomrule
    \end{tabular}
\end{table}

Results in Table~\ref{tab:data_quantity} show that increasing the amount of unlabeled data during pretraining improves downstream performance. The model pretrained on 60K samples achieves 2.0\% higher accuracy in the 10-shot regime and 0.8\% in the 50-shot regime compared to using only 10K samples. In the 1-shot setting, it is 3.6\% higher, but this difference is within one standard deviation. Notably, both pretraining configurations substantially outperform the supervised baseline in low-data regimes, with the performance gap narrowing as more labeled data is available for fine-tuning.

\section{Conclusion}

In many event-based applications, labels are the bottleneck and models must be adapted from only a few annotated samples. In this work, we presented EventCLR, a contrastive self-supervised approach to train Spiking Neural Networks for event-based vision. Its main contribution is the adaptation of SimCLR to event-based datasets, with a recipe to augment the input data, shown on a time-stepped SNN. Our results demonstrate that by leveraging event-specific augmentations, EventCLR learns robust and transferable representations from unlabeled event data, improving data efficiency in low-data regimes. We showed consistent gains over supervised baselines on CIFAR10-DVS, N-Caltech101, N-MNIST, DVS-Gesture, and N-Cars, particularly in few-shot and semi-supervised settings. Our ablation study showed that, used alone, temporal and polarity augmentations are more useful than spatial transformations, but that spatial and temporal augmentations are the most complementary pair, and the combination of the three families works best. Additionally, we found that computing the contrastive loss at each timestep independently yields downstream accuracy comparable to time-averaging embeddings before the loss, and we adopt the simpler time-averaged strategy throughout. Furthermore, we showed that learned features transfer across datasets, including datasets recorded with different sensors, and that on N-MNIST more unlabeled pretraining data (60K \vs 10K samples) improves the learned representations.

SimCLR-style contrastive learning is usually reported to require large batch sizes (or memory banks) to provide enough negatives, which is a memory bottleneck for spiking models trained with BPTT. In our experiments on CIFAR10-DVS, however, reducing the batch size from 256 to 8 did not degrade few-shot accuracy, which suggests that this constraint is less strict with long pretraining on small datasets. Whether this holds for larger and more diverse pretraining sets remains to be tested, and non-contrastive approaches (e.g., SimSiam, BYOL) could also be considered in that case. The temporal aspect is also only weakly tested: we use $T = 4$ time steps, and the comparison with a non-spiking ResNet-18 was done on CIFAR10-DVS only. Since the memory cost of BPTT grows with the number of time steps, the small batch sizes that work here could allow more time steps, and so a better assessment of the role of temporal dynamics, in particular on DVS-Gesture. We also did not measure energy consumption, synaptic operations, or execution on neuromorphic hardware, leaving the potential deployment benefits motivating this work to be evaluated in future studies. Finally, extending this self-supervised approach to other downstream tasks, such as object detection, depth estimation, or optical flow, will be crucial for improving the capabilities of event-based vision systems.

\ack{This work is under the programme DesCartes and is supported by the National Research Foundation, Prime Minister’s Office, Singapore, under its Campus for Research Excellence and Technological Enterprise (CREATE) programme. This work was also supported by the ANR grant ANR-11-LABX-0040 within the French State Programme “Investissements d’Avenir”. We thank the anonymous reviewers for their constructive comments.}

\section*{Use of generative AI}
Generative AI tools were used to help with writing and to generate parts of the code. They were not used for data analysis or to produce scientific content, and the authors checked all text and code.

\bibliographystyle{unsrt}
\bibliography{main}
\end{document}